\documentclass[letterpaper,10pt,conference]{support/ieeeconf}
\IEEEoverridecommandlockouts

\usepackage{amsmath,amssymb,amsfonts,bm}
\usepackage{graphicx}
\usepackage{booktabs,multirow,array}
\usepackage{url}
\usepackage{cite}
\usepackage[hidelinks]{hyperref}
\usepackage{xcolor}
\usepackage{threeparttable}
\usepackage{float}
\usepackage{subcaption}
\usepackage{makecell}

\newcommand{\norm}[1]{\left\lVert #1 \right\rVert}
\title{\LARGE \bf LV-Calib: LiDAR-Camera Extrinsic Calibration with Boundary-Response Modeling}

\author{Sheng~Hong
\thanks{Sheng Hong is with the Pen-Tung Sah Institute of Micro-Nano Science and Technology, Xiamen University, Xiamen 361102, China. E-mail: \texttt{shenghong@xmu.edu.cn}.}
}

\begin{document}
\maketitle
\thispagestyle{empty}
\pagestyle{empty}

\begin{abstract}
In this work, we present \emph{LV-Calib}, a calibration framework for
LiDAR-camera extrinsic estimation and LiDAR boundary-response calibration
using a printable planar target. The target serves as a shared observation
carrier: visual fiducials provide indexed image measurements, while circular
reflectivity boundaries provide LiDAR-observable structural feature points.

Instead of directly fitting boundary points as ideal geometric contours,
LV-Calib automatically crops background points, estimates the target plane,
and iteratively refines accurate LiDAR-side 3-D feature points from intensity
and geometric constraints. The refinement explicitly handles the broadened and
distorted transition band induced by finite beam footprint and mixed-intensity
returns, which is commonly observed around black-white reflectivity
discontinuities.
Given these refined LiDAR features, we formulate a weighted reprojection-consistent
extrinsic optimization with LiDAR feature alignment. Image observations are
kept in the reprojection domain to account for pixel-level measurement
uncertainty, while LiDAR feature alignment is weighted by refinement
confidence. This avoids over-confident 3-D alignment and reduces the
propagation of image-resolution noise into the extrinsic estimate.
Finally, using the estimated extrinsic and the extracted beam-footprint-induced transition band, LV-Calib calibrates the LiDAR boundary response by estimating pitch-yaw-range residual statistics of boundary-overlap samples, yielding a practical uncertainty model for LiDAR returns near reflectivity discontinuities.

Experiments on printed-board calibration data demonstrate sub-pixel
reprojection accuracy, millimeter-level LiDAR feature consistency, and improved
odometry performance. The calibrated extrinsic and boundary-response statistics
can further support uncertainty-aware LiDAR-visual SLAM and mapping. Code and
calibration data are available at
\href{https://github.com/sheng00125/LV-Calib}{\textcolor{blue}{https://github.com/sheng00125/LV-Calib}}.
\end{abstract}

\section{Introduction}

Recent years have witnessed a rapid growth of multi-sensor fusion in robotic perception, localization, mapping, and 3D reconstruction. LiDAR provides metric geometric measurements and remains reliable under illumination changes, while cameras provide dense visual appearance, texture, and semantic cues. IMUs further complement both sensors with high-rate motion constraints. Owing to these complementary properties, LiDAR-inertial and LiDAR-visual-inertial systems have become increasingly popular for robust state estimation and mapping, ranging from LiDAR-inertial odometry~\cite{shan2020lio,xu2021fast,xu2022fast} to tightly coupled LiDAR-visual-inertial pipelines such as LIC-Fusion, R2LIVE, LVI-SAM, R3LIVE, FAST-LIVO, FAST-LIVO2, and R$^3$LIVE++~\cite{zuo2019lic,lin2021r,shan2021lvi,lin2022r,zheng2022fast,zheng2024fast,lin2024r}. Recent systems further combine LiDAR-camera-inertial fusion with radiance-field or Gaussian-based mapping, enabling accurate colored point clouds, dense scene reconstruction, and real-time photorealistic map rendering~\cite{hong2024liv,hong2025gs}.
\begin{figure}[tp]
	\centering
	\includegraphics[width=1.05\linewidth]{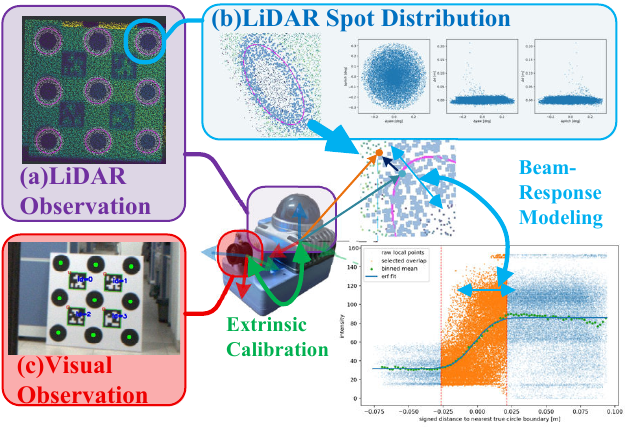}
\caption{LV-Calib: joint LiDAR-camera extrinsic and boundary-response calibration with a printable target.}
	\vspace{-2pt}
	\label{fig.hardware2}
\end{figure}

However, the effectiveness of these systems relies critically on accurate geometric relationships among sensors. In particular, the LiDAR-camera extrinsic calibration determines how 3D LiDAR measurements are projected into images and how visual observations are associated with metric geometry. Even small extrinsic errors may lead to biased colorization, inconsistent cross-modal residuals, degraded data association, and accumulated map distortion in tightly coupled odometry and reconstruction pipelines. This issue becomes more pronounced in practical self-assembled platforms, where LiDARs and cameras are frequently re-mounted, replaced, or roughly aligned from CAD models. Therefore, repeatable and easy-to-deploy LiDAR-camera extrinsic calibration remains an essential engineering and algorithmic problem.

Existing LiDAR-camera calibration methods are commonly divided into target-based and targetless categories. Targetless methods are attractive because they avoid dedicated calibration targets, but they usually rely on sufficient scene structure, texture, or semantics, and their robustness can drop in low-texture, repetitive, or weakly informative environments. Target-based methods are often more repeatable and easier to validate, but many recent high-accuracy methods for solid-state LiDARs depend on custom-made 3D objects, hollow structures, or CNC machining. In practice, this shifts part of the calibration burden from algorithm design to target fabrication and deployment.

A second issue is usually overlooked. When a LiDAR beam scans across a high-contrast black-white boundary, the returned points do not form an ideal infinitely sharp contour. Instead, the finite illuminated footprint, mixed returns near the boundary, detector integration, and view-dependent sampling geometry create a finite transition band. In practice, this may appear as boundary bleeding, streaking, or unstable edge points. Most calibration pipelines either suppress such points by robust fitting or absorb them into an empirical threshold. However, for high-precision LiDAR SLAM and map building, these effects matter: they perturb boundary localization during calibration, and they also reveal device-specific uncertainty that should not be ignored when boundary points are reused downstream.

These two issues motivate the central question of this paper: can a single, fully printable target support both robust LiDAR-camera extrinsic calibration and a practical calibration of LiDAR boundary-response behavior? We answer this question positively. We design a standard AprilGrid board with circular black nodes inserted in the inter-tag gaps. The AprilTags provide robust camera-side localization and indexing, while the circular black-white boundaries create structured reflectivity transitions that are useful to the LiDAR. The board is planar, printable, and easy to reproduce, without requiring perforation, hollow machining, or special 3D fabrication.

Based on this shared observation target, we propose \emph{LV-Calib}. The
AprilTag observations are used to guide the target localization and suppress
background points in the LiDAR scan, rather than being treated as direct
LiDAR-side geometric measurements. After the target plane is estimated, the
LiDAR front-end iteratively refines the 3-D structural feature points from
intensity and geometric constraints, explicitly considering the broadened
transition band induced by finite beam footprint and mixed-intensity returns.
For extrinsic estimation, we formulate a weighted reprojection-consistent extrinsic optimization with
LiDAR feature alignment, where image observations remain in the reprojection
domain and LiDAR feature residuals are weighted by refinement confidence.
Finally, using the calibrated extrinsic and the extracted transition band,
LV-Calib calibrates the LiDAR boundary response by estimating
pitch-yaw-range residual statistics of boundary-overlap samples.

The main contributions are summarized as follows:
\begin{itemize}
    \item An automatic LiDAR-side structural feature extraction front-end is
    proposed for printable planar targets. Using AprilTag-guided target
    localization, the method suppresses background points and iteratively
    refines accurate 3-D feature points from black-white reflectivity
    transitions under a CAD-constrained boundary model. The refinement
    explicitly handles the broadened transition band caused by finite beam
    footprint and mixed-intensity returns.

    \item A weighted reprojection-consistent LiDAR-camera extrinsic optimization is
formulated with LiDAR feature alignment. The optimization accounts for
    pixel-level visual measurement uncertainty in the reprojection domain and
    weights LiDAR feature residuals according to refinement confidence,
    reducing over-confident 3-D alignment in the extrinsic estimate.

    \item A LiDAR boundary-response calibration procedure is introduced by
    using the calibrated extrinsic and boundary-overlap samples to estimate
    pitch-yaw-range residual statistics, providing a practical uncertainty
    model for LiDAR returns near reflectivity discontinuities.

    \item Extensive experiments on printed-board calibration data demonstrate
    sub-pixel reprojection accuracy and millimeter-level LiDAR feature
    consistency. The code and calibration dataset will be released to support
    reproducible evaluation.
\end{itemize}

\section{Related Work}

\subsection{Target-based LiDAR-camera calibration}

Target-based calibration remains a reliable choice when high robustness and
accuracy are required. Classical methods usually use checkerboards, polygonal
planar boards, or point-line-plane correspondences to estimate the rigid
transformation between LiDAR and camera~\cite{geiger2012automatic,park2014polygonal,zhou2018automatic}.
These methods can achieve good performance when the target is clearly observed
by both modalities. However, their accuracy strongly depends on the target
design and on whether reliable LiDAR-side correspondences can be extracted
from sparse, anisotropic, and noisy LiDAR scans.

Recent target-based methods have further improved the calibration efficiency
for modern LiDAR sensors. For example, FAST-Calib extracts geometric features
from a designed target and efficiently estimates the extrinsic
parameters~\cite{zheng2025fastcalib}. However, many target-based pipelines
still treat the extracted LiDAR features as ideal geometric measurements. In
practice, LiDAR points near reflectivity boundaries are affected by finite
beam footprint, mixed-intensity returns, and range-angular measurement
uncertainty, which can bias the recovered 3-D correspondences. In addition,
pure 3-D registration or SVD-based alignment usually hides image measurement
uncertainty inside camera-side 3-D points, so pixel-level localization noise
and image-resolution effects are not explicitly reflected in the final
extrinsic optimization. In contrast, this work focuses on refining LiDAR-side
structural feature points from boundary transition regions and estimates the
final extrinsic through a weighted optimization with reprojection consistency.

\subsection{Targetless and scene-based calibration}

Targetless calibration methods avoid dedicated calibration targets by relying
on scene geometry, image edges, semantic cues, or information-theoretic
alignment. Representative methods include mutual-information-based
calibration~\cite{pandey2015automatic}, unified spatiotemporal calibration
across multiple sensors~\cite{furgale2013unified}, and direct targetless
LiDAR-camera registration toolboxes~\cite{koide2023general}. Pixel-level
calibration further estimates the extrinsic by aligning natural edges observed
by LiDAR and camera, and studies LiDAR measurement characteristics for more
accurate edge extraction~\cite{yuan2021pixel}. MFCalib extends this line of
work by using multiple types of edge features, including depth-continuous
edges, depth-discontinuous edges, and intensity-discontinuous edges on
planes~\cite{ye2024mfcalib}.

These methods are attractive for in-situ deployment, but their robustness is
often strongly coupled with the observed environment and the quality of
initialization. Low-texture scenes, repetitive structures, weak geometric
constraints, or poor initial poses may lead to unstable convergence. In
contrast, this paper focuses on controlled calibration with a low-cost
printable board, where repeatability, LiDAR-side feature quality, and
measurement uncertainty modeling are prioritized.

\subsection{Visual fiducial targets in practical calibration}

AprilTag-based boards are widely used because they provide robust indexing
under viewpoint changes, scale variation, and partial occlusion~\cite{wang2016apriltag2}.
In our framework, the AprilGrid provides reliable camera-side indexing and
guides the localization of the target region in the LiDAR scan for background
suppression. The final LiDAR-side feature points are not directly obtained
from the vision-estimated board pose. Instead, they are iteratively refined
from LiDAR intensity and geometric constraints under a shared CAD layout. This
design allows visual fiducials to provide an effective search prior, while
ensuring that the final LiDAR correspondences are still determined by LiDAR
measurements themselves.

\subsection{LiDAR boundary response and downstream uncertainty}

LiDAR points near reflectivity discontinuities are often regarded as unstable
measurements in calibration and registration. However, these points also
contain information about the sensor response near boundaries. In
tightly-coupled LiDAR, LiDAR-visual, and LiDAR-visual-inertial systems,
extrinsic calibration errors and unreliable boundary measurements can
propagate to odometry, mapping, point-cloud colorization, and 3-D
reconstruction~\cite{zuo2019lic,shan2021lvi,lin2022r,zheng2022fast,zheng2024fast,lin2024r,hong2024liv,hong2025gs}.
Therefore, when boundary measurements affect calibration accuracy and
downstream map quality, they should not be simply ignored.

Our method uses circular reflectivity boundaries not only for extrinsic
calibration, but also for LiDAR boundary-response calibration. After the
LiDAR-side circular features and the extrinsic are estimated,
boundary-overlap samples around the recovered circular transitions are
associated with their nearest ideal boundaries. Their residuals are then
evaluated in pitch-yaw-range space, yielding a practical uncertainty model for
LiDAR returns near reflectivity discontinuities. Compared with existing
target-based methods, this work emphasizes both LiDAR-side feature refinement
and finite-beam-footprint-induced uncertainty modeling. The circular
structural feature points are recovered through CAD-constrained iterative
boundary refinement, while the broadened transition band is explicitly modeled
and calibrated, rather than being ignored and allowed to degrade the extrinsic
estimation accuracy.

\section{Methodology}

\subsection{System overview and observations}

\begin{figure}[h]
    \centering
    \includegraphics[width=\linewidth]{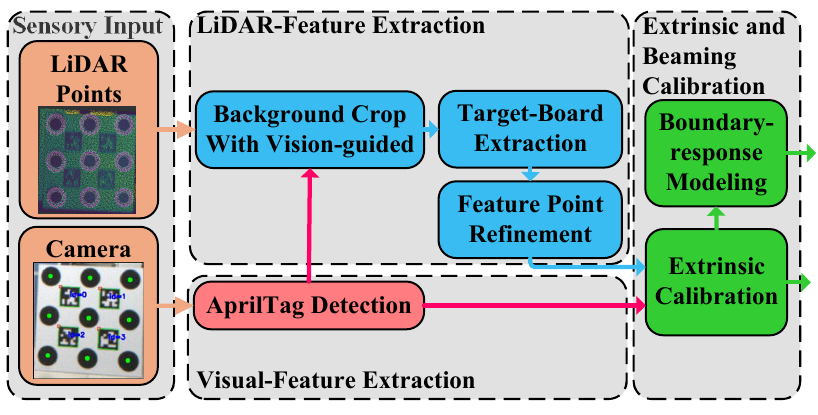}
    \caption{System overview of LV-Calib.}
    \label{fig:overview}
\end{figure}

Fig.~\ref{fig:overview} shows the overall pipeline of LV-Calib. The method
uses a printable planar target that serves as a shared observation carrier for
the camera and LiDAR. The AprilGrid provides indexed image measurements, while
the circular black nodes inserted in the inter-tag gaps provide
LiDAR-observable black-white reflectivity transitions. The visual observations
are mainly used to guide target localization and background suppression in the
LiDAR scan, which avoids blind search in 3-D space and improves the efficiency
of target-plane extraction. The final LiDAR-side feature points are not
directly taken from the vision-estimated board pose; instead, they are refined
from LiDAR intensity and geometry.

Let $\mathbf{p}_i^B$ be the designed circular-node center in the board frame
$\{B\}$, and let ${}^{C}\mathbf{T}_{L}$ denote the LiDAR-to-camera extrinsic
to be estimated. For scene $s$, the camera observes AprilTag corners
$\mathbf{z}_{s,m}\in\mathbb{R}^2$, and the LiDAR front-end extracts circular
feature centers $\mathbf{p}_{s,i}^{L}\in\mathbb{R}^3$ together with boundary
samples around each circle. The calibration problem is to recover a single
${}^{C}\mathbf{T}_{L}$ that explains all scenes while respecting the metric
layout of the printed target.

For each scene, AprilTag detections first provide a vision-guided target
localization prior. This prior is only used to crop a board-aligned LiDAR
region and remove most background points. A target plane is then estimated
from the cropped LiDAR points, and the local points are represented in a
board-plane coordinate system as $\mathbf{u}_j=[u_j,v_j]^\top$ with intensity
$I_j$. The initial 2-D node layout is obtained from the known CAD geometry and
is then refined by the iterative procedure described below.

\subsection{CAD-constrained iterative refinement of LiDAR circular features}
\label{subsec:angle_distance_refinement}

The key LiDAR-side problem is to recover accurate 3-D structural feature
points from noisy boundary measurements. Directly fitting the observed
boundary points as ideal circular contours is unstable, because LiDAR points
around black-white transitions are not located on a sharp physical edge. Due
to partial beam hits, mixed returns, and range-angular uncertainty, the
observed edge is often broadened, dragged, or locally distorted. Therefore,
LV-Calib estimates the circular features iteratively: local boundary
transitions provide center corrections, and the shared CAD layout regularizes
all circular nodes jointly.

For the $i$-th circular node with current center
$\mathbf{c}_i=[c_u,c_v]^\top$ and radius $r_i$, local LiDAR samples are first
unwrapped into angle--distance coordinates:
\begin{equation}
\begin{aligned}
\theta_j &= \mathrm{atan2}(v_j-c_v,u_j-c_u) \\
d_j &= \norm{\mathbf{u}_j-\mathbf{c}_i}-r_i
\end{aligned}
\label{eq:angle_distance_unwrap}
\end{equation}
where $\theta_j$ is the angular coordinate of the sample around the current
circle center, and $d_j$ is its signed radial distance to the current circular
boundary. This representation preserves the angular location of each sample.
A biased circle center does not only produce a constant radial offset;
instead, it appears as a low-frequency angular variation of the transition
midpoint. This property makes the center correction observable from the
boundary response.

The circle is divided into angular bins. In each valid bin, the inner and
outer intensity levels are estimated by robust medians, and the transition
midpoint $d_k^\star$ is obtained by interpolating the signed distance where
the intensity crosses the mid-level. A first-order perturbation of the circle
gives
\begin{equation}
d_k^\star \simeq
\Delta c_u\cos\theta_k+\Delta c_v\sin\theta_k+\Delta r
\label{eq:first_order_boundary_wave}
\end{equation}
Stacking all valid bins gives an over-determined linear system for
$\Delta\bm{\xi}_i=[\Delta c_u,\Delta c_v,\Delta r]^\top$, which is solved by
weighted least squares. The bin weights increase with intensity contrast and
sample support, while bins with insufficient support or weak contrast are
rejected. The update is step-limited to avoid fitting high-frequency noise in
the transition band.

The circular features are not optimized independently. After each local update
round, the refined 2-D centers $\tilde{\mathbf{c}}_i$ are projected back to a
single scale-fixed CAD layout:
\begin{equation}
\min_{\mathbf{R}_2\in\mathrm{SO}(2),\,\mathbf{t}_2}
\sum_i \alpha_i
\norm{\tilde{\mathbf{c}}_i-(\mathbf{R}_2\mathbf{q}_i^B+\mathbf{t}_2)}^2
\label{eq:layout_rigid_refinement}
\end{equation}
where $\mathbf{q}_i^B$ is the designed board-plane coordinate of the $i$-th
node. The scale is fixed by the printed board, and no Sim(2) or Sim(3)
rescaling is introduced. The updated CAD layout predicts the next-round
center positions, and the angle--distance refinement is repeated until the
center correction becomes small. Finally, the refined 2-D centers are lifted
back to the LiDAR frame using the estimated target plane.

\subsection{Weighted registration with reprojection consistency}

After LiDAR-side structural features are refined, the remaining problem is to
estimate the LiDAR-camera extrinsic from visual and LiDAR observations. A
common solution is to first recover camera-side 3-D nodes from the PnP board
pose and then perform 3-D registration. However, these reconstructed 3-D nodes
inherit finite image resolution, tag-corner localization noise, and fixed
intrinsic-calibration errors, which are not directly observed as isotropic
3-D noise.

LV-Calib therefore uses a joint weighted formulation that keeps the visual
constraint in the image domain and aligns the LiDAR nodes in the metric board
domain. The optimization jointly estimates the global extrinsic
${}^{C}\mathbf{T}_{L}$ and the per-scene board poses ${}^{C}\mathbf{T}_{B,s}$:
\begin{equation}
\begin{aligned}
\min_{{}^{C}\mathbf{T}_{L},\{{}^{C}\mathbf{T}_{B,s}\}}
&
\sum_{s,m}
\frac{
\left\|
\mathbf{z}_{s,m}-
\pi({}^{C}\mathbf{T}_{B,s}\mathbf{P}_m^B)
\right\|^2
}{\sigma_{\rm px}^2}
\\
&+
\sum_{s,i}
\frac{
\left\|
({}^{C}\mathbf{T}_{B,s})^{-1}
{}^{C}\mathbf{T}_{L}\mathbf{p}_{s,i}^{L}
-
\mathbf{p}_i^B
\right\|^2
}{\sigma_{s,i}^2}.
\end{aligned}
\label{eq:weighted_pnp_lidar}
\end{equation}
The first term constrains AprilTag corners by reprojection with pixel noise
scale $\sigma_{\rm px}$, while the second term constrains each refined LiDAR
node by its board-frame fitting uncertainty $\sigma_{s,i}$. Thus pixels and
meters are not added directly; both residuals are normalized into
dimensionless weighted errors. In our implementation, $\sigma_{s,i}$ is derived
from the LiDAR circle-refinement error with a minimum noise floor. The
separate 3-D-only, reprojection-only, and joint objectives in the experiments
further verify the contribution of each term.

\subsection{LiDAR boundary-response calibration}

The refined circular features also provide a natural reference for LiDAR
boundary-response calibration. Once the circular nodes and the extrinsic are
estimated, the corresponding ideal circular boundaries are known in the LiDAR
frame. This makes it possible to compare LiDAR points in the transition band
with their nearest ideal boundaries, instead of treating these points only as
unstable outliers.

For each LiDAR point near a recovered circular boundary, we compute its signed
distance $d$ to the nearest estimated circle. The intensity transition around
the boundary is modeled by
\begin{equation}
\begin{aligned}
I(d) &=
\mu_{\rm in}(1-\lambda(d))+\mu_{\rm out}\lambda(d) \\
\lambda(d) &=
\frac{1}{2}\left[
1+\mathrm{erf}\left(\frac{d-\delta}{\sqrt{2}\sigma}\right)
\right]
\end{aligned}
\label{eq:erf_transition}
\end{equation}
where $\mu_{\rm in}$ and $\mu_{\rm out}$ denote the two intensity levels, and
$\lambda(d)$ describes the smooth transition between them. The fitted
transition width is used only to select boundary-overlap samples within the
mixed-response region. It is not interpreted as a direct estimate of the
physical LiDAR beam diameter.

For each measured LiDAR point $\mathbf{x}^L$, we find the nearest ideal boundary
point $\hat{\mathbf{x}}^L$ and compare the two points in pitch-yaw-range
coordinates:
\begin{equation}
\Delta\mathbf{q} =
\left[
\phi(\mathbf{x}^{L})-\phi(\hat{\mathbf{x}}^{L}),\;
\psi(\mathbf{x}^{L})-\psi(\hat{\mathbf{x}}^{L}),\;
r(\mathbf{x}^{L})-r(\hat{\mathbf{x}}^{L})
\right]^{\top}
\label{eq:pyr_residual}
\end{equation}
The empirical mean and covariance of $\Delta\mathbf{q}$ form a practical
uncertainty model for LiDAR returns near reflectivity discontinuities. This
model characterizes the angular and range dispersion of boundary points around
an ideal reflectivity edge, and converts the transition band into calibrated
boundary-response statistics for downstream registration and mapping.

\section{Experiments}

\subsection{Equipment Setup}
As shown in Fig.~\ref{fig.hardware2}, our calibration setup consists of a LiDAR, a camera, and the proposed printable target. We evaluate the proposed LiDAR-side feature extraction and refinement front-end on three static calibration scenes, denoted as Scene A--C. In each scene, the same printed target is observed under a different board pose. The camera-side AprilTag detection is used to guide background cropping in the LiDAR scan and to more accurately isolate the board point cloud. Based on the cropped board region, the target plane is estimated and the nine circular reflectivity nodes are initialized from the known CAD layout. The same parameters are used for all scenes.

Our experiment focuses on the effectiveness of accurate LiDAR-side circular feature extraction, since these features directly determine the 3-D correspondences used in the subsequent extrinsic calibration. To better reveal the behavior of the proposed refinement process, we visualize the intermediate angle--distance responses and the resulting center updates for each scene and each circular node.

\begin{figure}[t]
    \centering
    \includegraphics[width=\linewidth]{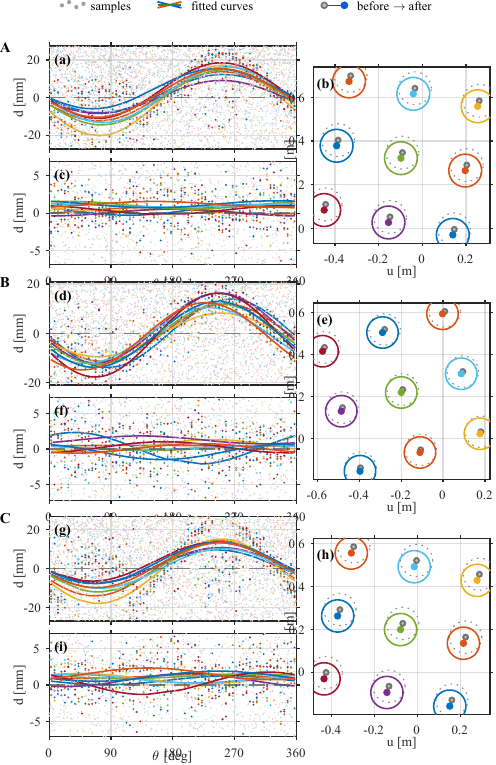}
\caption{Optimization process of the proposed method in three representative scenes.
(a)--(c) Scene A, (d)--(f) Scene B, and (g)--(i) Scene C.
For each scene, the top-left panel shows the initial angle--distance profile,
the bottom-left panel shows the refined profile, and the right panel shows
the corresponding circle-center updates on the calibration board.}
    \label{fig:feature_refinement_process}
\end{figure}

\subsection{LiDAR Feature Extraction and Refinement}
Fig.~\ref{fig:feature_refinement_process} summarizes the full refinement
process for all three scenes. The unwrapped angle--distance representation
preserves the angular position of each LiDAR sample around a circular node.
Therefore, a center bias appears as a structured low-frequency variation in
the boundary response rather than being averaged out by a one-dimensional
radial profile. This makes the update direction directly observable from the
shape of the fitted response.

Before refinement, the fitted responses show scene-dependent deviations caused
by the initial CAD transfer, point sparsity, and mixed returns around the
reflectivity boundary. After refinement, the responses become more centered
around the nominal boundary and the remaining fluctuations are largely
symmetric around zero. This indicates that the first-order center error has
been removed while avoiding an over-sensitive fit to local noise.

The right column of Fig.~\ref{fig:feature_refinement_process} further shows
that the node updates are not solved independently. The final centers are
constrained by the same scale-fixed CAD layout, so the nine nodes move
coherently as one board-level structure. This is important for calibration:
the extracted LiDAR features remain geometrically consistent with the printed
target, while still exploiting the local intensity transition around each
node. Across the three scenes, the visualization confirms that the proposed
front-end can produce stable LiDAR-side 3-D node correspondences from a
standard printed board without relying on a special mechanical target.

\subsection{Extrinsic calibration accuracy}

Table~\ref{tab:extrinsic_method_comparison} compares several extrinsic
estimation objectives on the same three-scene dataset. The compared methods
cover three types of formulations. 3D-SVD performs closed-form 3-D alignment
of cross-modal nodes, as commonly used in target-based calibration
pipelines~\cite{zheng2025fastcalib}. W-SVD further introduces measurement
weighting in the 3-D alignment stage, but still operates purely in the
reconstructed 3-D domain. PnP-reproj. mainly optimizes the image reprojection
error, which is related to image-plane refinement strategies used in prior
calibration methods~\cite{yuan2021pixel}. In contrast, the proposed weighted
reprojection-consistent formulation keeps the visual constraints in the
AprilTag-corner reprojection domain while aligning the refined LiDAR nodes in
the metric board domain. Therefore, both visual measurement uncertainty and
LiDAR feature uncertainty are explicitly reflected in the optimization.

The AD-ref. rows use the proposed CAD-constrained iterative refinement,
whereas the raw rows directly use the initialized LiDAR node centers. Even for
closed-form 3-D alignment, AD-refinement improves the result from 2.324 to
2.073 px in projection RMS and from 27.620 to 26.564 mm in LiDAR-node
residual, confirming the benefit of accurate LiDAR-side circular feature
recovery. More importantly, the proposed formulation achieves substantially
better metric consistency than the 3-D alignment baselines. Compared with the
AD-refined 3D-SVD baseline, Ours with AD-ref. reduces the LiDAR-node residual
from 26.564 mm to 3.580 mm while maintaining a 0.973 px AprilTag-corner
reprojection RMS. Compared with W-SVD, the improvement indicates that
uncertainty weighting in the 3-D domain alone is insufficient when the
camera-side 3-D nodes already contain image measurement noise.

\begin{figure}[tp]
	\centering
	\includegraphics[width=1.0\linewidth]{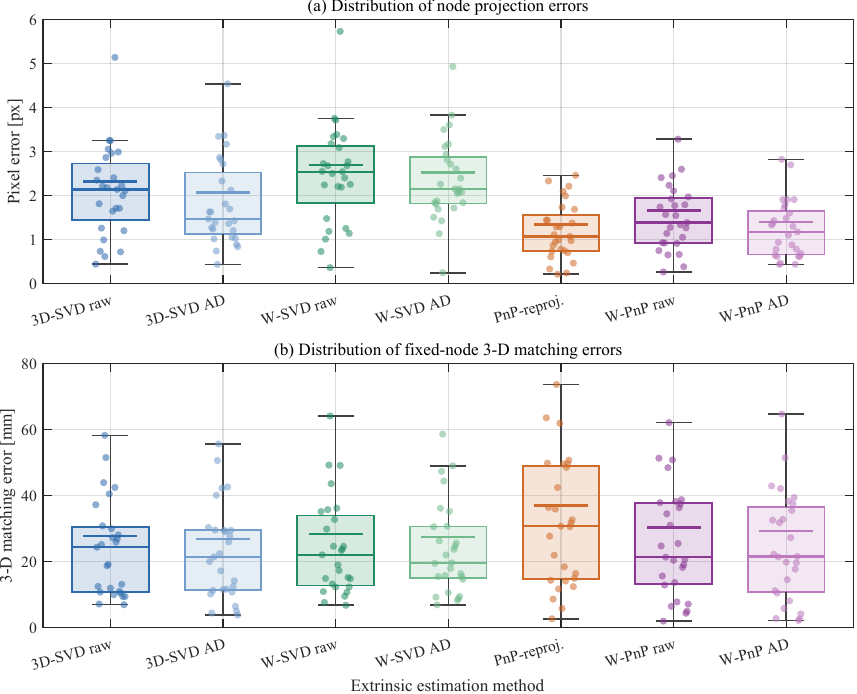}
	\caption{Residual distributions of different extrinsic estimation objectives. Ours with AD-ref. achieves low errors in both the image and 3-D domains.}
	\vspace{-2pt}
	\label{fig.extrinsic_method_residual_boxplot}
\end{figure}

Although PnP-reproj. obtains a relatively low image-plane error, its 3-D
consistency degrades noticeably, suggesting that a purely image-domain
objective can absorb depth uncertainty and overfit the extrinsic estimate. In
contrast, the proposed formulation uses the board layout as a geometric prior
and jointly optimizes reprojection consistency and LiDAR metric alignment,
making it less sensitive to noisy reconstructed 3-D visual correspondences.
Fig.~\ref{fig.extrinsic_method_residual_boxplot} further visualizes the
residual distributions in both the image and 3-D domains. Ours with AD-ref.
achieves low reprojection error and low 3-D alignment error simultaneously,
whereas the competing methods tend to favor only one domain. Therefore, we use
Ours with AD-ref. as the final extrinsic estimate.

\begin{table}[t]
\centering
\caption{Comparison of camera--LiDAR extrinsic estimation objectives.}
\label{tab:extrinsic_method_comparison}
\setlength{\tabcolsep}{3.0pt}
\renewcommand{\arraystretch}{1.05}
\scriptsize
\begin{tabular}{@{}llcrr@{}}
\toprule
Method & Centers & Visual term & RMS$_{\rm px}$ & RMS$_{\rm mm}$ \\
\midrule
3D-SVD~\cite{zheng2025fastcalib} & raw     & node   & 2.324 & 27.620 \\
3D-SVD~\cite{zheng2025fastcalib} & AD-ref. & node   & 2.073 & 26.564 \\
W-SVD                            & raw     & node   & 2.695 & 26.468 \\
W-SVD                            & AD-ref. & node   & 2.528 & 25.537 \\
PnP-reproj.~\cite{yuan2021pixel} & raw     & node   & 1.346 & 36.892 \\
Ours                             & raw     & corner & 0.972 & 4.866 \\
Ours                             & AD-ref. & corner & 0.973 & 3.580 \\
\bottomrule
\end{tabular}
\end{table}

\subsection{Boundary-response calibration results}

We further use the extracted overlap region together with the recovered CAD
circle centers to calibrate the LiDAR boundary response. In this stage, the
fitted black--white transition in Eq.~\eqref{eq:erf_transition} is only used
to localize the mixed-response region near each circular boundary. After the
overlap region is identified, each measured LiDAR point is paired with its
nearest estimated true boundary point, and the residual is evaluated in the
LiDAR pitch--yaw--range coordinates defined in Eq.~\eqref{eq:pyr_residual}.

\begin{figure}[t]
    \centering
    \includegraphics[width=\linewidth]{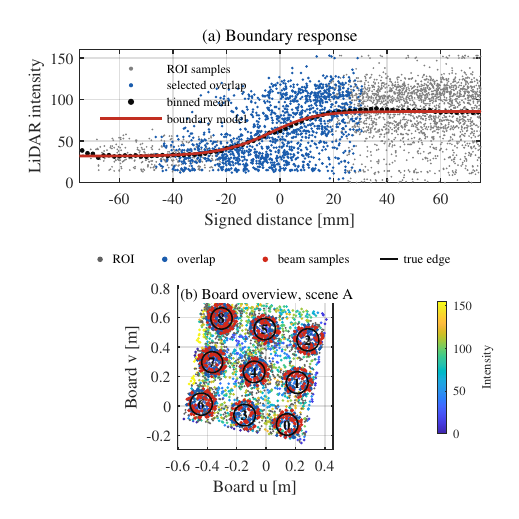}
    \caption{LiDAR boundary-response calibration using black--white circular
    boundaries. Top: LiDAR intensity as a function of signed boundary
    distance, showing the boundary ROI, the overlap region near the mixed
    response, and the fitted transition model. Bottom: board-frame
    visualization of the LiDAR intensity map, boundary ROI, selected overlap
    region, calibration points, and recovered circular boundaries.}
    \label{fig:beam_response_calibration}
\end{figure}

Fig.~\ref{fig:beam_response_calibration} first verifies the quality of this
calibration process. The top panel aggregates the LiDAR intensity responses
from the three scenes. The gray points indicate the boundary ROI, while the
selected points around the black--white transition correspond to the overlap
region used for boundary-response calibration. Their binned means are well
described by the error-function model. The bottom panel provides a board-frame
visualization for a representative scene. The boundary ROI, selected overlap
region, calibration points, and recovered circular boundaries are shown
together in the board plane. It can be seen that the selected calibration
points consistently follow the nine circular boundaries, while the recovered
true boundaries remain aligned with the designed target layout. This confirms
that the calibration is spatially tied to the intended reflectivity
discontinuities rather than to arbitrary intensity measurements.

\begin{table}[t]
\centering
\caption{Pitch--yaw--range residual statistics of the calibrated LiDAR boundary response.}
\label{tab:beam_response_stats}
\setlength{\tabcolsep}{2.2pt}
\renewcommand{\arraystretch}{1.05}
\scriptsize
\begin{tabular}{@{}lrrrrrrr@{}}
\toprule
Scene & $N$ [k] & $\bar{\Delta\phi}$ & $\bar{\Delta\psi}$ & $\bar{\Delta r}$ & $\sigma_{\phi}$ & $\sigma_{\psi}$ & $\sigma_r$ \\
& & [deg] & [deg] & [mm] & [deg] & [deg] & [mm] \\
\midrule
A & 28.0 & 0.001 & 0.002 & 1.44 & 0.165 & 0.165 & 9.38 \\
B & 18.9 & 0.001 & 0.002 & -0.77 & 0.177 & 0.178 & 10.23 \\
C & 16.0 & 0.001 & -0.001 & -0.14 & 0.100 & 0.100 & 7.05 \\
All & 63.0 & 0.001 & 0.001 & 0.38 & 0.155 & 0.156 & 9.17 \\
\bottomrule
\end{tabular}
\end{table}

Table~\ref{tab:beam_response_stats} reports the calibrated residual
statistics over the three scenes. The mean residuals in pitch, yaw, and range
are all close to zero, with overall values of only 0.001 deg, 0.001 deg, and
0.38 mm, respectively. This indicates that, after nearest-boundary pairing,
the selected overlap points are not systematically displaced to one side of
the recovered circular boundaries. The calibrated standard deviations are
0.155 deg, 0.156 deg, and 9.17 mm in pitch, yaw, and range, respectively.
Notably, the pitch and yaw dispersions are of similar magnitude, whereas the
range residual exhibits a larger spread. This is consistent with the different
error characteristics of angular and range measurements in Livox
non-repetitive scanning LiDARs. Scene C shows the tightest distribution, with
approximately 0.100 deg angular standard deviation and 7.05 mm range standard
deviation, whereas Scenes A and B are wider.

\begin{figure}[t]
    \centering
    \includegraphics[width=\linewidth]{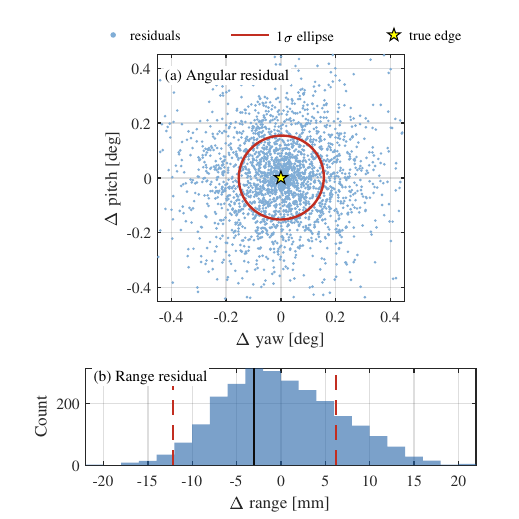}
    \caption{Pitch--yaw--range residual distribution of the calibrated
    boundary-response points. The angular residuals form a compact footprint
    around the recovered true boundary, while the range histogram is centered
    near its empirical peak with the one-standard-deviation interval marked by
    dashed lines.}
    \label{fig:beam_response_pyd_uncertainty}
\end{figure}

Fig.~\ref{fig:beam_response_pyd_uncertainty} further visualizes the overall
residual distribution. The angular residuals are concentrated around the
recovered true boundary, and the one-standard-deviation ellipse summarizes
their lateral spread in the pitch--yaw plane. The range histogram is also
compact around its empirical peak, with most selected points lying within the
calibrated $\pm\sigma_r$ interval. Overall, these results statistically characterize the measurement uncertainty of the LiDAR boundary response in pitch--yaw--range space.

\section{Conclusion}

We presented \emph{LV-Calib}, a calibration framework for joint LiDAR-camera
extrinsic estimation and LiDAR boundary-response calibration using a printable
planar target. The method uses AprilTag-guided target localization to suppress
background points, iteratively refines LiDAR-side circular features under a
CAD-constrained model, and estimates the final extrinsic with a weighted
reprojection-consistent formulation. The same recovered boundaries are further
used to estimate pitch-yaw-range residual statistics for LiDAR
boundary-response calibration. Experiments demonstrate sub-pixel reprojection
accuracy, millimeter-level LiDAR feature consistency, and improved odometry
performance. These results show that LV-Calib provides a practical calibration
procedure for accurate cross-modal alignment and LiDAR boundary-response
modeling.

\end{document}